\documentclass[letterpaper, 10pt, conference]{ieeeconf}      %

\IEEEoverridecommandlockouts                              %

\let\labelindent\relax                                    %

\usepackage{graphicx}            %
\usepackage{color}
\usepackage{amsmath}             %
\usepackage{amssymb}             %
\usepackage{amsfonts}            %
\usepackage{enumitem}            %
\usepackage{multirow}            %
\usepackage{siunitx}             %
\usepackage{url}                 %
\usepackage{xspace}              %
\usepackage[T1]{fontenc}         %
\usepackage[hidelinks]{hyperref} %
\usepackage{balance}
\usepackage[bottom]{footmisc} 
\usepackage{xfrac}
\usepackage{tensor}
\usepackage{todonotes}

\usepackage[firstpage=true]{background}
\newcommand\copyrighttext{%
\parbox{\textwidth}{
\footnotesize
}}

\SetBgContents{\copyrighttext}
\SetBgScale{1}
\SetBgColor{black}
\SetBgAngle{0}
\SetBgOpacity{1}
\SetBgPosition{current page.north}
\SetBgVshift{-0.8cm}

\usepackage[bottom]{footmisc}

\graphicspath{{images/}}
\DeclareGraphicsExtensions{.pdf,.png,.jpg,.jpeg}

\newcommand{\T}{\mathbb{T}} %
\newcommand{\mypm}{\mathbin{\mathpalette\@mypm\relax}}

\makeatletter
\newcommand{\amsray}{%
\mathpalette {\overarrow@\rayfill@}}
\def\rayfill@{\arrowfill@{\mkern4mu\mapstochar\relbar}\relbar{\mkern 4.08mu}}%
\makeatother

\newcommand{\seclabel}[1]{\label{sec:#1}}

\newcommand{\figlabel}[1]{\label{fig:#1}}
\newcommand{\tablabel}[1]{\label{tab:#1}}
\newcommand{\eqnlabel}[1]{\label{eqn:#1}}

\newcommand{\secref}[1]{Section~\ref{sec:#1}\xspace}

\newcommand{\figref}[1]{Fig.~\ref{fig:#1}\xspace}
\newcommand{\tabref}[1]{Table~\ref{tab:#1}\xspace}
\newcommand{\eqnref}[1]{(\ref{eqn:#1})\xspace}

\title{\LARGE \textbf{Centroidal State Estimation and Control\\for Hardware-constrained Humanoid Robots}}
\author{Grzegorz Ficht and Sven Behnke%
\thanks{All authors are with the Autonomous Intelligent Systems (AIS) Group, Computer Science Institute VI,
        University of Bonn, Germany. Email: {\tt\small ficht@ais.uni-bonn.de}.}}

\usepackage{eso-pic}

\AtBeginDocument{\AddToShipoutPictureFG*{\AtTextUpperLeft{\put(0,\LenToUnit{9pt}){\parbox{\textwidth}{\centering\bfseries
IEEE-RAS 22nd International Conference on Humanoid Robots (Humanoids), Austin, USA, December 2023.
}}}}}
\begin{document}

\bstctlcite{IEEEexample:BSTcontrol}

\maketitle
\thispagestyle{empty}
\pagestyle{empty}

\begin{abstract}

We introduce novel methods for state estimation, feedforward and feedback control, which specifically target humanoid 
robots with hardware limitations. Our method combines a five-mass model with approximate dynamics of each mass. 
It enables acquiring an accurate assessment of the centroidal state and Center of Pressure, 
even when direct forms of force or contact sensing are unavailable. Upon this, we develop a feedforward scheme that operates on the 
centroidal state, accounting for insufficient joint tracking capabilities. Finally, we implement feedback mechanisms, 
which compensate for the lack in Degrees of Freedom that our NimbRo-OP2X robot has. The whole approach allows for reactive stepping 
to maintain balance despite these limitations, which was verified 
on hardware during RoboCup 2023, in Bordeaux, France. 

\end{abstract}
\section{Introduction}

Robust bipedal locomotion remains a coveted research goal, still posing several challenges despite much work in the field. 
The difficulties stem not only from the high dimensionality of the problem, underactuation, and complex dynamics, but also 
originate from practical issues such as robot hardware limitations. 

If real-time operation and the gap between models and reality wouldn't be an issue, optimization-based methods could be used to plan motions 
capable of exploiting complex dynamics~\cite{al2012trajectory}\cite{lengagne2013generation}. Reduced-order models provide 
tractability by focusing on the core dynamics of the system, e.g. planning and executing 
Center of Mass~(CoM) trajectories given a sequence of Center of Pressure~(CoP) locations~(footholds). Examples include Linear 
Quadratic Regulation~(LQR)~\cite{kajita2010biped} and Model Predictive Control~\cite{dimitrov2011sparse}. The simplified dynamics
often allow for finding closed-form solutions, such as for stabilizing the Zero Moment Point~(ZMP)~\cite{tedrake2015closed}. 
The problem can also be further reduced by splitting the CoM dynamics into stable and unstable components, where only the unstable 
ones need to be handled. The unstable part is usually called the Instantaneous Capture Point~\cite{pratt2006capture}, or its 3D 
generalization: The Divergent Component of Motion~\cite{englsberger2011bipedal}\cite{englsberger2015three}. Single mass models 
can also be extended with a flywheel, which models the Centroidal Angular Momentum~(CAM) generated around the CoM~\cite{pratt2006capture}. 
In recent years, the concept of Centroidal Dynamics~\cite{orin2013centroidal} is becoming more popular, allowing for expressive 
upper body movement  and CAM regulation~\cite{dai2014whole}. In this regard, there are still open questions, such as 
finding a meaningful~\textit{angular equivalent} to CoM~\cite{chen2022angular} or generating reference CAM trajectories~\cite{schuller2021online}. 
Our most recent work~\cite{ficht2023direct} falls into the CoM-ZMP category, but also tackles the mentioned issues related to CAM, 
where we propose to use the orientation of the Principal Axes of Inertia for direct CAM regulation.

A common issue of adopting ZMP-based approaches in physical robots is the assumption that the joint trajectories producing stable 
motions can be accurately tracked. In lower quality hardware, where torque is limited, latencies are significant, and backlash is 
present, this is not the case. Our humanoid NimbRo-OP2X~\cite{ficht2018nimbro} falls into this category, and---so far---was 
able to walk only by using hand-crafted gaits with Central Pattern Generation~(CPG)~\cite{Allgeuer2016a}\cite{missura2019capture}. 
To achieve better tracking performance, feedforward techniques are typically applied on the joint level. These methods often require 
some system and parameter identification and runtime computation of expected loads~\cite{Schwarz2013a}. The modelling complexity of 
such schemes can approach extreme levels~\cite{gehring2016practice}, but learning-based approaches can alleviate the need to deal 
directly with these intricacies~\cite{hwangbo2019learning}. The robot itself might also not be capable of such dynamic movement, though
high-bandwidth Quasi-Direct Drive~(QDD) actuator technology currently used in quadrupeds allows for shaping ground reaction forces 
within milliseconds~\cite{wensing2017proprioceptive}. Feedforward techniques remain an important component for bipedal walking, 
at least until humanoids utilizing QDD actuation become more wide-spread~\cite{chignoli2021humanoid}.

\begin{figure}[!t]
\centering{\includegraphics[width=0.99\linewidth]{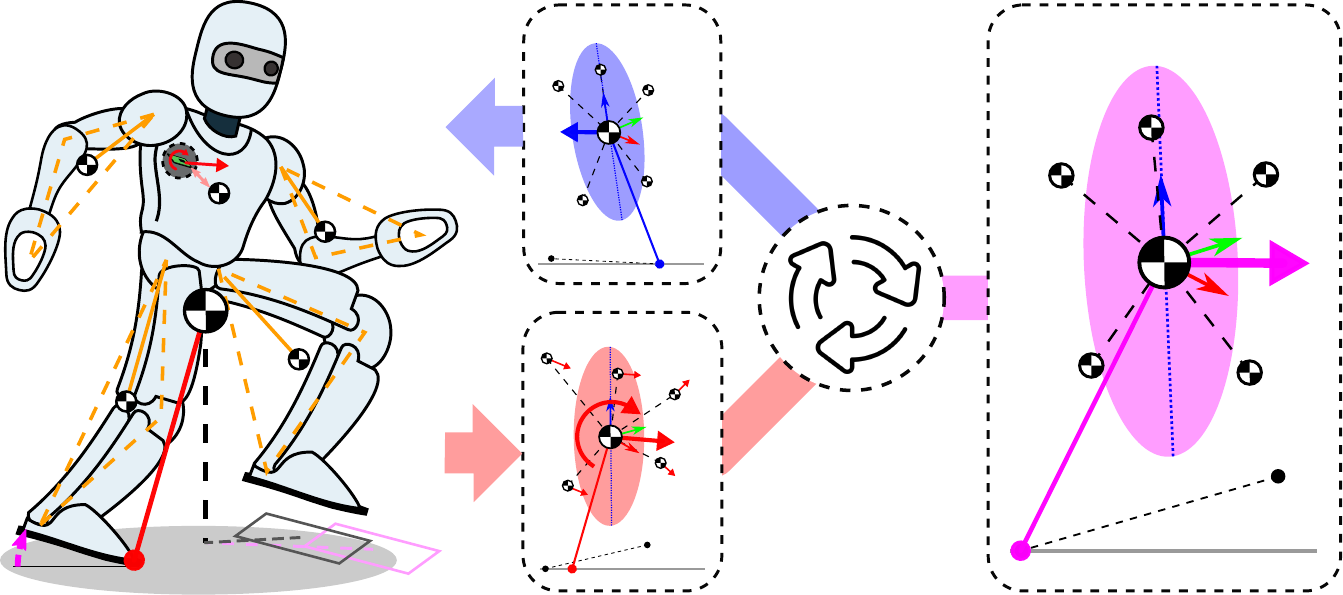}}
\caption{Centroidal Control of humanoid robots with hardware limitations. The IMU readings and limb dynamics are fused into a centroidal state of the robot. Three of such models form the basis of a feedforward compensator. Reactive stepping feedback is used to counter tilting due to the robot's kinematic limitations.}
\figlabel{inertia_teaser} 
\vspace{-2ex}
\end{figure}

Another requirement for ZMP control is to actually sense the CoP, which is done using Force-Torque~(FT) sensors. FT sensors are 
versatile in the sense that they measure not only if the limb is in contact with the environment, but what wrench is acting on the system through it. 
However, they suffer from noise and creep due to the constant forces they are subjected to. In systems like ours, without any direct 
form of contact and force sensing, estimating the external wrench working on the it and assuring balance is not trivial. Current non-FT ZMP 
estimators still rely on having Force Sensitive Resistors~(FSRs), or similarly to our previous work~\cite{ficht2023direct} assume 
the measured torso acceleration approximates the CoM acceleration~\cite{masuya2020review}. In this paper, we show that a higher-accuracy 
estimate can be achieved by including limb dynamics, even without foot sensors altogether. Aside from restricted sensing options, 
our system is also limited in its kinematic structure. Due to a double 4-bar linkage parallel kinematics design, the legs have 
only 5 Degrees of Freedom~(DoF) and leave the torso-foot pitch angle locked to be orthogonal. This makes the robot prone to tilting, and poses an 
additional challenge to the balance controller.

The contributions of this paper can be split into three main components,  (see \figref{inertia_teaser}):
\begin{itemize} 
\item an accurate centroidal state estimator that uses a five-mass model incorporating limb dynamics; 
\item a simple, yet effective feedforward scheme, compensating for state tracking inaccuracies by operating on the centroidal state; and 
\item hardware-focused correction mechanisms for a robot with kinematic restrictions, extending previous work~\cite{ficht2023direct}.
\end{itemize} Ultimately, we show that even limited hardware can achieve relatively high performance in bipedal walking with adequate state estimation and control.

\section{Five-Mass Model} \seclabel{model_reduction}

The following section summarizes our five-mass representation, describing how single mass kinematics and dynamics relate to joint angles and overall system properties.

\subsection{Limb CoM} \seclabel{limbcom}

The Center of Mass~(CoM) for each limb is described using a triangle parameterization, denoted by $P = (p_s,p_l)\in[0,1]\subset\mathbb{R}$. The \textit{side distribution} $p_s$ divides the lower link of the limb, forming a vector between the division point and the limb's origin. The \textit{length distribution} $p_l$ represents the ratio of the centroid's distance to said vectors length. For example, a triangle with uniform density has $p_s = \frac{1}{2}$, $p_l = \frac{2}{3}$, and the origin to CoM vector lies on a median (see \figref{triapprox}).

Both the kinematics and dynamics of a limb's mass $\mathbf{m}_*$ are predominantly determined by the limb extension and rotation. The weight of the end effector (hand, foot) is included with adequate $P$ values, but its orientation is omitted due to the negligible effect it has on $\mathbf{m}_*$. Hence, only the first $n$ limb joints $\mathbf{q}_*$ are necessary to obtain $\mathbf{m}_*$; with the limb transform $\textbf{T}_*$ tying these values through $P$:
\begin{equation} \eqnlabel{limb_mass}
\begin{aligned}
\textbf{m}_* &= \textbf{o}_* + \textbf{T}_* &, \\
\textbf{T}_* = \begin{bmatrix}T_x\\T_y\\T_z\end{bmatrix}=\textbf{R}_\text{O} &\begin{bmatrix}-ap_lp_s \sin(q_n) \\0\\-(a p_l p_s \cos(q_n) + c p_l)\end{bmatrix} &,\\
\end{aligned}
\end{equation}
where $q_n$ is the bending joint angle (knee, elbow), $a,c$ are triangle sides, and $\textbf{R}_\text{O}$ is a matrix obtained from a sequence of $n$$-$$1$ rotations performed at the limb origin $\textbf{o}_*$ (hip, shoulder). We can then compute the limb mass Jacobian $\textbf{J}_*$:

\begin{equation} \eqnlabel{jacobian}
\textbf{J}_* = \begin{bmatrix} \frac{\partial T_x}{\partial q_1} \dots \frac{\partial T_x}{\partial q_n}\\
\frac{\partial T_y}{\partial q_1} \dots \frac{\partial T_x}{\partial q_n}\\
\frac{\partial T_z}{\partial q_1} \dots \frac{\partial T_x}{\partial q_n}\end{bmatrix}\,;\quad\dot{\textbf{m}}_* = \textbf{J}_*\dot{\textbf{q}}_*
\end{equation}
tying the limb joint velocities $\dot{\textbf{q}}$ to its CoM velocity $\dot{\textbf{m}}_*$, which will play an important role in~\secref{estimation}.

\begin{figure}[!t]
\centering{\includegraphics[width=0.99\linewidth]{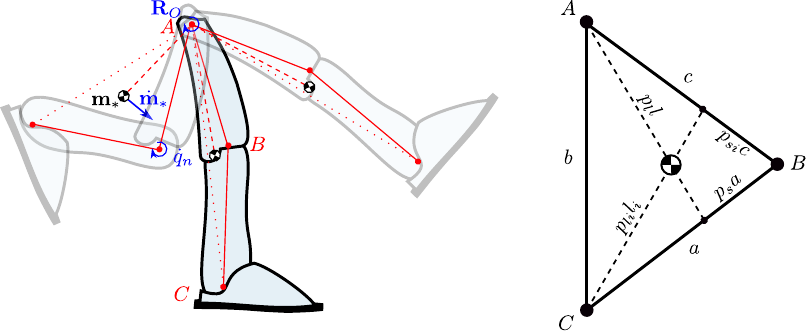}} %
\caption{Limb kinematics and dynamics in relation to mass positioning. Using a triangle mass distribution---parametrized in $P$---a direct relation between limb mass movement and joint angles is achieved.}
\figlabel{triapprox}
\vspace{-2ex}
\end{figure}

\subsection{System CoM and Inertia}

The root of the robot is defined as a floating 6 DoF base frame  $\textbf{\textit{F}}_{B}$ attached to the torso with an orientation of $\textbf{R}_t$ and its position centered between the hip joint origins $\textbf{h}_m$. The head---due to its limited range of motion---is considered as a part of the torso, with their combined mass position $\textbf{m}_t$ set at a configurable offset from  $\textbf{\textit{F}}_{B}$.  This leaves its movement unconstrained, which is advantageous for visual tracking algorithms. The robot's limbs originate at an offset $\textbf{o}_*$ to the torso and their joint configuration results in the (left and right) leg $\textbf{m}_{ll}$, $\textbf{m}_{rl}$ and arm $\textbf{m}_{la}$, $\textbf{m}_{ra}$ CoM (\secref{limbcom}).
Combining these five masses $\textbf{m}_i$ and accounting for their physical weight $w_i$ we achieve the overall CoM position $\textbf{m}_c$:
\begin{equation} \eqnlabel{totalmass}
\textbf{m}_c = \frac{ \sum_{i}^{}(\textbf{m}_i w_i)}{\sum_{i}^{}w_i}\,;\quad i\in\{t,ll,rl,la,ra\}.
\end{equation} 

The five masses also shape the robot's inertia $\textbf{I}_R$, which we partition into the principal axis form using the moments $\textbf{I}_{PA}$ and orientation $\textbf{R}_I$:
\begin{equation}\eqnlabel{principalmoments}%
\begin{aligned}
\mathbf{I}_R = \mathbf{R}_I\mathbf{I}_{PA}\mathbf{R}_I^\top = \mathbf{R}_I\begin{bmatrix}I_{xx} &0 &0\\0 &I_{yy} &0\\0 &0 &I_{zz}\end{bmatrix}\mathbf{R}_I^\top.
\end{aligned}
\end{equation}
The leg masses form a dumbbell which rotates around their aggregate mass $\textbf{m}_l$ named the \textit{lower body} mass. This is repeated for the \textit{upper body}, where the arm masses are combined with half of the torso mass each into $\textbf{m}_{lu}$, $\textbf{m}_{ru}$ and rotate around $\textbf{m}_u$. The vector from $\textbf{m}_{u}$ to $\textbf{m}_{l}$ passes through $\textbf{m}_{c}$ and its normal sets the \textit{tilt} z-axis $\textbf{z}_I$ of the orientation $\mathbf{R}_I$. The definition is complete with the \textit{yawing} component $\psi_I$, computed from weighted yaw angles of the upper $\psi_u$ and lower dumbbells $\psi_l$, around $\textbf{z}_I$:
\begin{equation}\eqnlabel{yawangles}%
\begin{gathered}
\psi_I = \frac{\psi_uw_u+\psi_lw_l}{w_u+w_l},\\
w_u = w_{t}+w_{ra}+w_{la},\qquad w_l=w_{ll}+w_{rl}.
\end{gathered}
\end{equation}
The computation of $\mathbf{R}_I$ from $\textbf{z}_I$ and $\psi_I$ follows the intermediate tilt representation from~\cite{allgeuer2015fused}.

Our five-mass representation describes the physical properties of the system with high fidelity across a wide range 
of configurations. The tunable mass parameters can account for even large modelling errors and be fitted to the actual robot, 
greatly simplifying the hardware transfer. As the mapping between our five-mass representation and kinematics is unique, we developed an inertia-shaping 
whole-body inverse kinematics technique in~\cite{ficht2020fast}. By providing the foot-relative desired centroidal properties $\textbf{m}_c$, $\textbf{R}_I$, 
and $\textbf{I}_{PA}$ over time, we achieve simultaneous control over linear and angular momenta in a uniform fashion.
The splitting of the inertia tensor into the Principal Axes form and the defined, continuous behavior of $\textbf{R}_I$ as an \textit{angular counterpart} to the CoM
is what sets our approach apart from typically used methods, e.g. the CoM Jacobian.

\section{State Estimation}  \seclabel{estimation}

The robot's state can be captured through various quantities, some of which can be observed directly with physical sensors, others indirectly using soft sensors. The Inertial Measurement Unit~(IMU) is crucial to our approach. It is a versatile sensor providing not only the floating base (torso) orientation $\textbf{R}_t$~\cite{Allgeuer2014}~\cite{ludwig2018comparison}~\cite{cisneros2020lyapunov}, but also the external forces acting on it through accelerations. The pose of the robot is reconstructed by measuring the joint angles \textbf{q} with encoders, which also measure $\dot{\textbf{q}}$. Next, foot frames $\textbf{\textit{F}}_{FR}$, $\textbf{\textit{F}}_{FL}$ are obtained, where the lower one is assumed to be the supporting leg. To prevent unwanted toggling of support exchanges, a threshold on the height difference is set. With the whole pose reconstructed, we compute the centroidal properties $\textbf{m}_c$ and $\textbf{I}_R$. %

\subsection{Linear Motion} %

Considering the system as a point mass simplifies analyzing the forces acting on it, which can be directly inferred through accelerations. In previous work~\cite{ficht2023direct}, we used a Kalman Filter~(KF) to combine $\textbf{m}_c$ and sensed torso acceleration $\ddot{\textbf{r}}_s$, with the assumption that $\ddot{\textbf{r}}_s \approx \ddot{\textbf{m}}_t \approx \ddot{\textbf{m}}_c$. The torso acceleration is rotated from its own frame~$^T\ddot{\textbf{m}}_t$, into a global $\ddot{\textbf{m}}_t$, with gravity $\textbf{g}$ subtracted: 
\begin{equation} \eqnlabel{globaltrunkacc}
\ddot{\textbf{m}}_t = {^T\ddot{\textbf{m}}_t} \textbf{R}_t - \textbf{g}.
\end{equation}
This provides fast and sufficiently precise linear motion state estimates, when the CAM is close to zero. This simple method is insufficient, though, when the torso is rotating or when walking at higher speeds---due to significant limb swinging. By addressing these two issues, we can reduce the reliance on this assumption. Firstly, the IMU sensor mounting $\textbf{r}_s$ rarely coincides with $\textbf{m}_t$. Thus, angular accelerations of the torso $\ddot{\boldsymbol{\theta}}_t$ contribute a tangential component to the linear ${^T\ddot{\textbf{r}}_s}$ measurement:

\begin{equation} \eqnlabel{tangentialacc}
{^T\ddot{\textbf{r}}_{s}} = {^T\ddot{\textbf{m}}_t} + \ddot{\boldsymbol{\theta}}_t \times (\textbf{r}_s - \textbf{m}_t).
\end{equation}
To obtain $\ddot{\boldsymbol{\theta}}_t$, we simply perform finite numerical differentiation on the gyro measurements. The amplified noise from the differentiation is handled by increasing the corresponding measurement covariance matrix values of the KF.
By rearranging and incorporating~\eqnref{tangentialacc} into~\eqnref{globaltrunkacc}, we remove the ${\ddot{\textbf{r}}_s}\approx{\ddot{\textbf{m}}_t}$ assumption and achieve a purely linear $\ddot{\textbf{m}}_t$.

The second issue refers to the correlation between ${\ddot{\textbf{m}}_t}$ and ${\ddot{\textbf{m}}_c}$, in light of the mass distribution. The torso is the root link anchoring the limbs and their movement will generate reaction forces around their origin. These forces displace the torso, similarly to external ones which the IMU detects through ${\ddot{\textbf{r}}_s}$ and ${\dot{\boldsymbol{\theta}}_t}$ as disturbances. We incorporate limb dynamics by considering each limb's~\textit{global} acceleration ${\ddot{\textbf{m}}_{*}}$---a combination of the origin's absolute and origin-relative mass accelerations:

\begin{equation} \eqnlabel{limbglobacc}
{\ddot{\textbf{m}}_{*}} = {\ddot{\textbf{m}}_t} + \ddot{\boldsymbol{\theta}}_t \times (\textbf{o}_* - \textbf{m}_t) + {^T\ddot{\textbf{m}}_*} \textbf{R}_t .
\end{equation}
The total mass acceleration $\ddot{\textbf{m}}_c$ is then achieved by weighing the five-mass accelerations similarly to~\eqnref{totalmass}. This is then supplied into the KF instead of $\ddot{\textbf{m}}_t$, thus removing the ${\ddot{\textbf{m}}_t}\approx{\ddot{\textbf{m}}_c}$ assumption. The origin-relative limb mass accelerations ${^T\ddot{\textbf{m}}_*}$ were obtained through differentiation of ${^T\dot{\textbf{m}}_*}$ from~\eqnref{jacobian}. 

\subsection{Angular Motion}

Concurrent to linear motion, the contribution of angular motion is equally important in shaping the system dynamics. Control over it necessitates estimating the angular state. We have briefly introduced it in~\cite{ficht2023direct}, with the state being reconstructed purely from Euler angles $\boldsymbol{\theta}_I$ encoded in the inertia orientation $\textbf{R}_I$. There was a significant delay associated with this approach though, which rendered the estimates inoperable. Specifically, before the state of angular accelerations finally converged, the real accelerations could have already significantly changed. Similarly to the linear motion estimation, we can improve on this by combining the inertia angles $\boldsymbol{\theta}_I$ with the angular accelerations of the five masses around $\textbf{m}_c$ (denoted as  ${^c\ddot{\boldsymbol{\theta}}_i}$) in the measurement model of the KF. 
For each of the masses $\ddot{\textbf{m}}_i$, it is important to consider that CAM is generated only if they differ from $\ddot{\textbf{m}}_c$~\cite{herr2008angular}.
The accelerations of the five masses ${^c\ddot{\boldsymbol{\theta}}_{i}}$ are then:
\begin{equation} \eqnlabel{limbangacc}
{^c\ddot{\boldsymbol{\theta}}_{i}} =\frac{ (\textbf{m}_i-\textbf{m}_c)\times({\ddot{\textbf{m}}_i} - \ddot{\textbf{m}}_c )}{\|(\textbf{m}_i-\textbf{m}_c) \| ^2}.\\
\end{equation}
Again, simply weighing these angular accelerations as in~\eqnref{totalmass} yields the total system centroidal acceleration ${^c\ddot{\boldsymbol{\theta}}_I}$. 
Alternatively, the angular motion could be calculated using velocities. As $\dot{\boldsymbol{\theta}}_t$ along with all limb velocities $\dot{\textbf{m}}_*$ are readily available, the only requirement would be to estimate $\dot{\textbf{m}}_t$. We decided against this as it would require chaining filters, potentially introducing stability issues.

\section{Control} \seclabel{control}

The following section describes all the control components specific to the presented approach. As the methods presented are an extension of the authors previous work~\cite{ficht2023direct}, it is recommended to use it as reference for a better overview.

\subsection{Reference Trajectories}

The open-loop trajectory generation scheme is similar to the one previously presented in the DCC~\cite{ficht2023direct}, albeit with modifications and improvements. The Center of Mass $\textbf{m}_c$ and its dynamics resolve with time $t$ from two planar ($\textbf{m}_{c,x},\textbf{m}_{c,y}$) Linear Inverted Pendulum Models~(LIPM) with a height of $\textbf{m}_{c,z}=h$ and Center of Pressure (CoP)~${\textbf{r}}_{p}$~\cite{kajita20013d}:
\begin{equation} \eqnlabel{lipm}
\begin{gathered} 
\ddot{\textbf{m}}_{c,x} = \omega^2({\textbf{m}}_{c,x}-{\textbf{r}}_{p,x}), \quad \omega = \sqrt{\mathbf{g}_z/h},\\
{\textbf{m}}_{c,x}(t) = {\textbf{m}}_{c,x_0}\cosh(\omega t)+\dot{\textbf{m}}_{c,x_0}\frac{1}{\omega}\sinh(\omega t),\\ 
\dot{\textbf{m}}_{c,x}(t) = {\textbf{m}}_{c,x_0}\omega\sinh(\omega t)+\dot{\textbf{m}}_{c,x_0}\cosh(\omega t).\\
\end{gathered} 
\end{equation}

Walking velocity $\textbf{v}_g$ is controlled by setting a constant gait frequency $f_g$ and altering the stride length. Step placement $\textbf{s}$ is initially selected to satisfy the symmetry constraint, but later refined to maintain continuous trajectories~\cite{kajita2019linear}. Previously, the swing foot trajectory would linearly progress from the last foothold to next one according to the step progression $\mu$. With the accumulated joint compliance and tracking errors, often the swing foot would contact the ground prematurely and still try to move. This degraded performance, introduced instabilities, and was harmful to the hardware. We alter the trajectory progression to $\mu_{a}$ with a customizable sigmoid $\sigma\left(\mu,p,m\right)$: %
\begin{equation} \eqnlabel{gaitprogression}
\begin{gathered} 
\mu_{a} =\frac{\sigma\left(\mu\right)-\sigma\left(0\right)}{\sigma\left(1\right)-\sigma\left(0\right)},\,\, \sigma\left(\mu,p,m\right)=\frac{1}{1+e^{p\left(m+0.5-\mu\right)}},\\
\mu\in[0,1),\qquad\qquad\qquad \mu[n+1] = \mu[n] + f_g \Delta t.
\end{gathered} 
\end{equation}
While $m\in(-1,1)$ shifts the midpoint of the output, $p$ adjusts the steepness of the sigmoid: $p\to0$ results in a linear progression and $p\to\infty$ in a step function.
The combination of these parameters allows for a speed-up in the progression and a settling period at the beginning of a step, its end, or both. 

To finalize the trajectory, the inertial properties $\mathbf{I}_R$ need to be set. The principal moments $\textbf{I}_{PA}$ are kept constant from a desired 
halt-pose, which are then usually adjusted by the whole body pose generator to satisfy other constraints ($\textbf{m}_c$, $\textbf{R}_I$). The robot's inertia 
orientation $\textbf{R}_I$ is set by combining a desired tilt z-axis $\textbf{z}_I$ and yaw angle $\psi_I$ (aligned with the step yaw progression). 
To prevent tilting moments around the CoM, $\textbf{z}_I$ is aligned with the robot tilt axis, giving a \emph{zero CAM heuristic}. In \cite{ficht2020fast},
the robot tilt axis was based purely on averaged foot frame position vectors $\textbf{r}_{FR}$, $\textbf{r}_{FL}$ pointed 
towards the CoM. With a kinematically locked torso, shaping the inertia is severely limited and might force the arms to overcompensate with undesired configurations
or rapid movement. We circumvent that by including $^T\textbf{m}_t$ and mass distribution in the calculation, thus finalizing $\textbf{R}_I$:
\begin{equation} \eqnlabel{inertiaref}
\begin{gathered}
\mathbf{z}_I = \frac{w_u}{\Sigma w_i}({^T\textbf{m}}_t)-\frac{w_l}{2\Sigma w_i}(\mathbf{r}_{FL}+\mathbf{r}_{FR}).
\end{gathered}
\end{equation}

\subsection{Feedforward Compensation}
\begin{figure}[!t]
\centering{\includegraphics[width=0.99\linewidth]{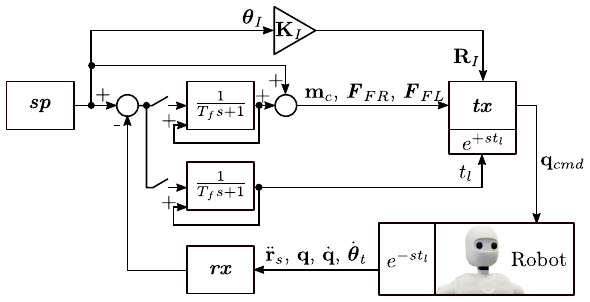}}
\vspace{-3ex}
\caption{Scheme of the feedforward compensation. The \textbf{\textit{sp}} model provides nominal setpoint trajectories, which get transmitted to the \textbf{\textit{tx}} model. Once the Lowpass Filters get enabled, they will start comparing the \textbf{\textit{sp}} values to the received ones in \textbf{\textit{rx}}, which will make the \textbf{\textit{tx}} diverge not only in values, but also forward it in time. This in turn brings \textbf{\textit{rx}} closer to \textbf{\textit{sp}}. }
\figlabel{feedforwardscheme}
\vspace{-3ex}
\end{figure}

To account for known disturbances, we apply a feed-forward control scheme. Unlike typically used methods which offset the position or provide expected 
torques to controllers in joint space, our approach focuses on the quantities relevant for the centroidal dynamics. 
While walking, a leg can either be supporting or swinging, which is tightly connected to the loads it is subjected to and thus how fast it can move. 
Factors such as external disturbances, joint friction, backlash, compliance, torque limitations, and sensorimotor delays can all contribute to poor 
state tracking. In the centroidal space, we identified and correct for the following cases:
\begin{itemize}
    \item \textbf{CoM to foot vector:} Both with respect to the support and swing foot, we track the $x,y$ values of $\textbf{m}_c - \textbf{r}_{FL/FR}$ to assure that steady-state offsets $\mathbf{r}^{ff}_{sup}$, $\mathbf{r}^{ff}_{sw}$ are generated and compensate for the torque necessary to carry the robot's mass, or place the swing foot in the desired position. 
    \item \textbf{CoM and step height:} Similarly to the $x,y$ displacement, from $\textbf{m}_c,z - \textbf{r}_{FL/FR,z}$ we infer if either of the feet need to be pushed down or pulled up to accommodate for carrying the respective body or leg weight.
    \item \textbf{CoM velocity:} Due to dynamically changing loads within the gait cycle as well as wrongly placed feet, the CoM might accumulate a velocity error $\dot{\textbf{m}_c}$ that needs to be taken into consideration by applying $\mathbf{r}^{v}$. 
    \item \textbf{Swing foot orientation:} The orientation of the swing foot $\textbf{R}_{FR/FL}$ is equally important to its placement, as it allows for a smooth support exchange to take place.
    \item \textbf{Support exchange event:} If the moment when the foot hits the ground varies from the expected one, then there exists a sensorimotor delay (latency) to account for. We quantify it by measuring the time difference $t_l$.
    \item \textbf{Inertia orientation:} By measuring the Euler angles $\boldsymbol{\theta}_I$ of $\textbf{R}_I$, it is easily observable if the upper body has sufficient torque to move as requested. Gains $\textbf{K}_I$ are applied to the set angles of the inertia rotation to either enforce more or decrease the rotation of the upper body.
\end{itemize}

The feedforward compensation scheme consists of three model states: \textbf{\textit{rx}}, \textbf{\textit{sp}}, \textbf{\textit{tx}} and a block of 
Lowpass Filters~(LPF), as shown in \figref{feedforwardscheme}. During walking, robot measurements of the above-mentioned quantities are put into 
\textbf{\textit{rx}} and compared against 
setpoint values \textbf{\textit{sp}} with the result placed in the respective value LPF. Due to the symmetry in the lateral movement, all $y$ values and 
the roll foot angle are unified with a multiplication by the leg sign $\iota$: $\{\text{L}=-1,\text{R}=1\}$. Toggling the filter input switch effectively 
enables/disables the feedforward term estimation process. Once the filters have been enabled, their output value will slowly approach the necessary 
compensation value. At the same time, the corrective terms are applied to the \textbf{\textit{tx}} model and transmitted to the robot, decreasing the 
$\textbf{\textit{sp}}-\textbf{\textit{rx}}$ errors. Before applying the corrections, the \textbf{\textit{tx}} trajectories are forwarded in time by the 
estimated $t_l$, which essentially predicts away the latency. The feedback loop in the LPF effectively turns them into integrators, keeping their value 
even if their input is disabled. The LPFs time constant $T_f$ approximately describes the necessary time for the feedforward terms to settle; we set it 
to a duration of several steps. The compensation variables are estimated at zero and nonzero walking velocities. Based on the desired walking speed, 
they are linearly interpolated for uniform operation.

The usage of the feedforward term depends on the controlled quantity. Unlike the straightforward application of the leg height and foot angles, 
the $x,y$ placement of the feet is less trivial. The trajectories need to be continuous between steps, meaning that they need to tolerate coordinate 
frame changes and smoothly fade out. The constant support $\mathbf{r}^{c}_{sup}$ and swing $\mathbf{r}^{c}_{sw}$ offsets, and combined with the velocity 
offset $\mathbf{r}^{v}$ into complete feedforward terms $\mathbf{r}^{ff}_{sup}$, $\mathbf{r}^{ff}_{sw}$ applied to the feet:
\begin{equation} \eqnlabel{feedforward}
\begin{gathered}
\mathbf{r}^{v}  = \left[\frac{\mu-0.5}{f_g}v_x\,,\, 0\,,\, 0 \,\right], \\
\mathbf{r}^{ff}_{sup} = \mathbf{r}^{c}_{sup} -  \mathbf{r}^{v},\\
\mathbf{r}^{ff}_{sw} = \mathbf{r}^{c}_{sw} + \mathbf{r}^{c}_{sup} + \mathbf{r}^{v},
\end{gathered}
\end{equation}
where $v_x$ is the estimated sagittal velocity offset. In the lateral direction, the foot placement was sufficient in achieving the desired velocity. 
As the swing foot can move more freely and is also carried by the support foot, we include $\mathbf{r}^{c}_{sup}$ in its applied feedforward vector 
$\mathbf{r}^{ff}_{sw}$. %
Finally, the feedforward terms are applied to the left and right foot, depending on the supporting leg $\iota_{tx}$.

\subsection{CoM Feedback Control} \seclabel{feedback}

Feedback control is assured with a CoM-ZMP controller initially proposed by Choi et al.~\cite{choi2007posture} and extended by us in~\cite{ficht2023direct}.
The controller realizes input-to-state stability by adjusting $\dot{\textbf{m}}_c$ to steer $\textbf{m}_c$ back to follow the nominal trajectories in the 
presence of errors, while considering the CoP position $\textbf{r}_p$. We modify our controller further by using the Centroidal Moment Pivot~(CMP) 
instead of ZMP, and include $\textbf{m}_c$ rate limiters to avoid large, abrupt CoM changes. 

\subsection{Tilt-compensating Step Feedback}

\begin{figure}[!t]
\centering{\includegraphics[width=0.99\linewidth]{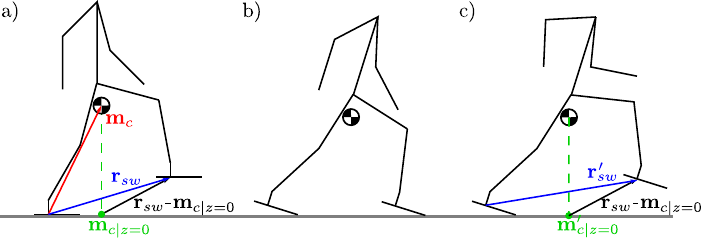}} %
\caption{Visual representation of the tilt compensation mechanism. a)~Nominal trajectory execution. b)~Preemptive touchdown due to unexpected tilt. c)~Corrective mechanism assuring approximate desired touchdown is achieved.}
\figlabel{tiltcomp}
\vspace{-1ex}
\end{figure}

By locking the foot pitch DoF, our robot is prone to tilting. Sufficiently large disturbances that are not naturally damped or dispersed by the feedback controller will create a torque in the sagittal direction and make the robot pivot about the edge of its feet. Depending on the combination of tilt and walking direction, the foot will either prematurely hit the ground at the wrong position or stay mid-air without performing a support exchange. Both of these situations result in quick loss of stability. To account for this, we transform the swing foot trajectory and aim for the swing foot to contact the ground at the correct $x$-distance from the CoM. We do so by: taking the nominal swing foot distance to the zero-tilt ground-projected CoM: $\textbf{r}_{sw}-\textbf{m}_{c|z=0}$, applying it to the tilt-rotated nominal CoM projection $\mathbf{m}'_{c|z=0}$, unrotating the resulting $\mathbf{r}'_{sw}$ and setting it as the commanded swing foot placement (see \figref{tiltcomp}). As a result, the step should place the CoM onto a trajectory close to the nominal one. In combination with the CoM controller, this is sufficient for the robot to tolerate moderate disturbances. For greater disruptions in the gait, we not only transform the foot trajectory but also recalculate the desired step placement. We do so by plugging the current state $\textbf{m}_c$, $\dot{\textbf{m}}_c$ and the remaining step time $\frac{1-\mu}{f_g}$ into the LIPM equations~\eqnref{lipm} to predict the End Of Step~(EOS) state $\textbf{m}^{eos}_c$ and compute its expected EOS error $\mathbf{e}^{eos}_c$. Instead of purely using the prediction, we linearly interpolate between the EOS error and the current CoM tracking error $\mathbf{e}_c = \mathbf{m}^{ref}_c - \mathbf{m}_c$ according to time passed. Using the symmetry assumption~\cite{kajita2019linear} we offset the next nominal step:
\begin{equation} \eqnlabel{feedforward}
\mathbf{s}\left[k+1\right]^{new} = \mathbf{s}\left[k+1\right] + 2(1-\mu)\mathbf{e}_c + 2\mu\mathbf{e}^{eos}_c.
\end{equation}

This assures that predictions made directly after a support exchange (when the foot is still aligning to the ground and CMP might fluctuate) do not falsely steer the gait. Naturally, once these errors reduce so does the commanded velocity and the robot returns to the nominal walking velocity. 

\section{Experimental Results}

The proposed approach is verified with a 135cm tall, NimbRo-OP2X humanoid~\cite{ficht2018nimbro} with 18 Degrees of Freedom. For ground-truth comparisons of 
estimates and quantitative results, we use a full MuJoCo simulation of the robot that also models the parallel linkages. The whole motion generation, 
estimation and control framework is running on-board the robot's computer, with a frequency of \SI{100}{Hz}.

\subsection{State Estimation}

\begin{figure}[!t]
\parbox{\linewidth}{\centering
\includegraphics[width=0.49\linewidth]{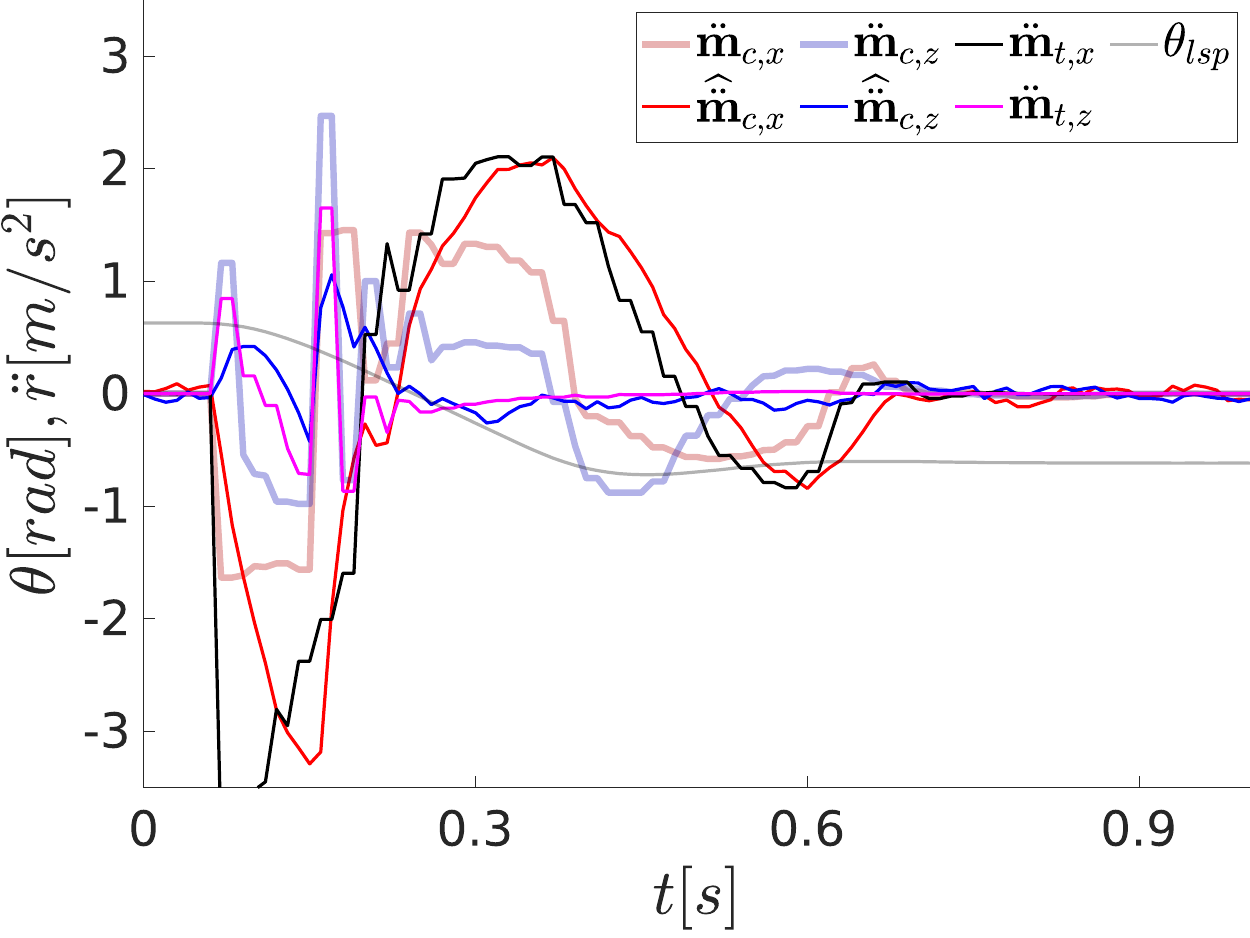}
\includegraphics[width=0.49\linewidth]{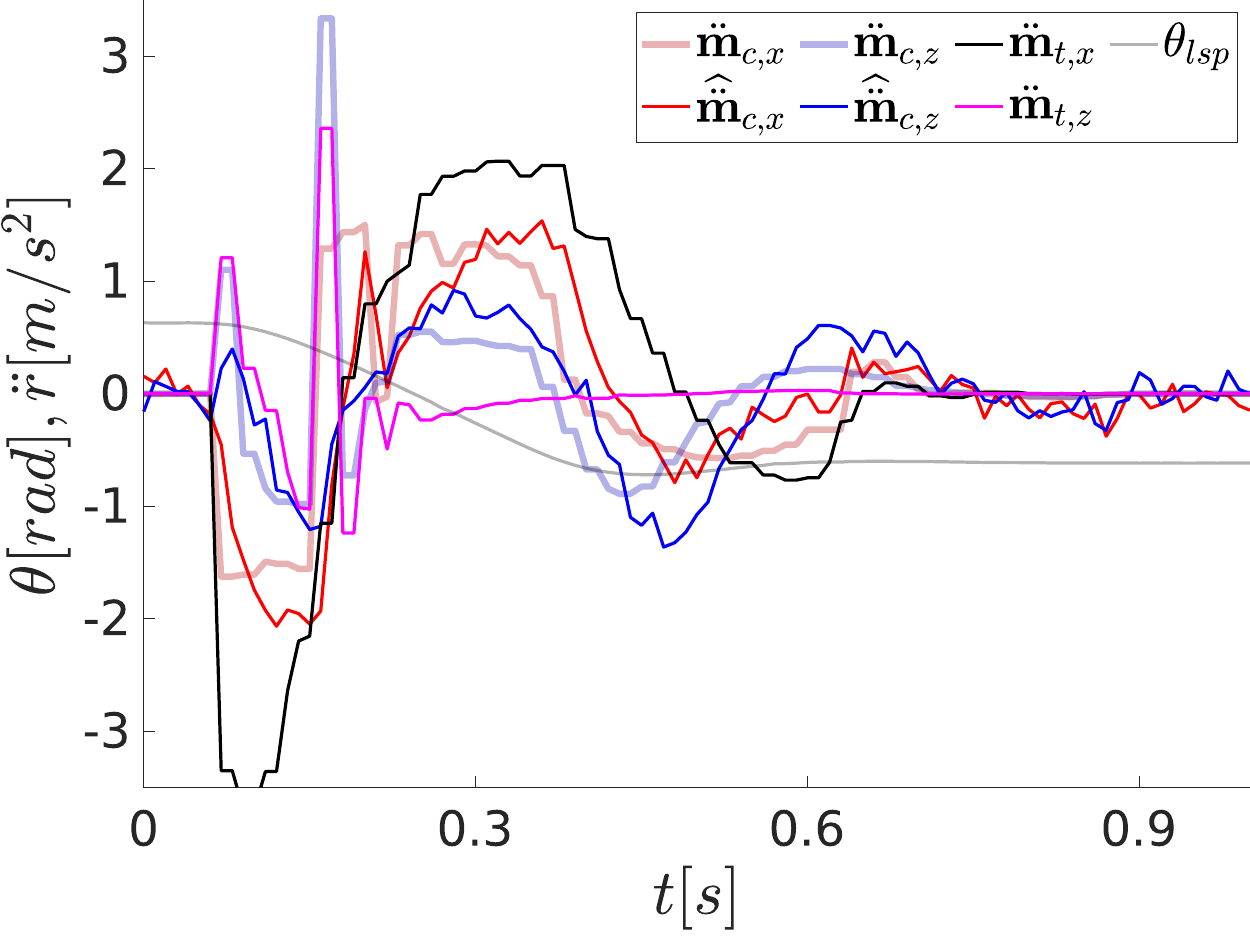}}
\caption{Comparison of a purely IMU-based estimator (left), and one which includes limb dynamics (right). The latter allows to capture the actual CoM dynamics, instead of the torso's. During both experiments, the robot is shifting its torso back and rotating fully extended arms forward. Left shoulder pitch angle, ground-truth torso and CoM accelerations along with their estimates are shown.}
\figlabel{estimation_accuracy}
\vspace{-2.5ex}
\end{figure}

To display the contribution of the limb dynamics in state estimation accuracy, the robot is tasked to slightly pitch its inertia $\textbf{R}_I$ backwards while
keeping $\textbf{m}_c$ still.
As the pitch angle of the torso is kinematically locked, the pose generator shifts the hip backward and rotates the fully extended arms towards the front to 
satisfy the constraints. The result of this experiment comparing the ground-truth and estimates of relevant variables, is shown in \figref{estimation_accuracy}.
As shown in~\cite{masuya2020review}, assuming that the IMU and CoM accelerations are synonymous is currently the most relevant approach when no force or 
torque measurements are available. Our previous estimator falls into this category and will serve as a baseline for further comparisons. 
Although it reacts swiftly to the movement, it does not capture $\ddot{\textbf{m}}_c$ well. It is also quite natural that it barely perceives any changes 
in $\ddot{\textbf{m}}_{c,z}$, as the torso just translates its position. The arms by rotating around 
their shoulder joints, change their mass position on the z-axis over time. This slightly shifts the aggregate mass of the robot in the z-direction, 
which is clearly observable with the presented approach. The newly included limb accelerations accurately offset the torso measurements and provide estimates 
visibly closer to $\ddot{\textbf{m}}_{c}$.

\begin{table}[!b]
\small
\centering
\setlength{\tabcolsep}{4pt}
\renewcommand{\arraystretch}{1.2}
\caption{State estimation error in simulation}
\begin{tabular}{l*{6}{c}}
\hline

\hline
& $e(\ddot{\textbf{m}}_{c,x})$ & $e(\ddot{\textbf{m}}_{c,y})$ & $e(\ddot{\textbf{m}}_{c,z})$ & $e(\dot{\textbf{m}}_{c,x})$ & $e(\dot{\textbf{m}}_{c,y})$ & $e(\dot{\textbf{m}}_{c,z})$ \\ 
\hline%
\hline%
\multicolumn{7}{c}{\scriptsize{Purely IMU-based estimation:}}\\ 
\hline%
$\mathbb{M}(e)$ & -0.5500 & -0.4070 & 0.0767 & -0.1146 & 0.0288 & 0.0424 \\
SD$(e)$ & 2.6092 & 1.3079 & \bf{4.2658} & 0.1039 & \bf{0.0356} & \bf{0.0829} \\
$\frac{\Sigma\left|e\right|}{T}$ & 1.8023 & 0.6752 & \bf{1.8965} & 0.1208 & 0.0367 & 0.0710 \\
\hline %
\multicolumn{7}{c}{\scriptsize{Including limb dynamics:}}\\
\hline
$\mathbb{M}(e)$ & \bf{0.0097} & \bf{-0.4010} & \bf{0.0540} & \bf{-0.0913} & \bf{0.0148} & \bf{0.0286} \\
SD$(e)$ & \bf{2.0921} & \bf{1.0489} & 5.5886 & \bf{0.0902} & 0.0358 & 0.0892 \\
$\frac{\Sigma\left|e\right|}{T}$ & \bf{1.1526} & \bf{0.6542} & 2.0439 & \bf{0.1058} & \bf{0.0269} & \bf{0.0671} \\

\hline

\hline
\tablabel{estimation_table}
\vspace{-2.5ex}
\end{tabular}
\end{table}  
We also verify the estimator in a more dynamic setting, where the simulated robot is made to walk forward open-loop at a constant velocity for 10 seconds. 
During this, the ground-truth and estimate values were gathered for the 3D CoM velocities and accelerations. From these, we computed the median 
$\mathbb{M}(e)$, standard deviation SD$(e)$ and average absolute $\frac{\Sigma\left|e\right|}{T}$ of the estimation errors. The results of these calculations 
were placed in \tabref{estimation_table}, with the top showing the previous, purely IMU-based estimator (baseline), and the bottom having the limb dynamics included.
A decrease in the estimation errors is clearly visible, which is most apparent in $\ddot{\textbf{m}}_{c,x}$. With all of the limbs swinging back and forth, their significant contribution to 
the overall system dynamics is simply omitted by the old method (as also shown in \figref{estimation_accuracy}). An interesting result arose with the 
$\ddot{\textbf{m}}_{c,z}$ estimate. Although the filter is able to capture more detail in the z-direction, it exhibits a higher standard deviation and 
average absolute error. The reason for this was made clear upon further inspection of the data. Increasing the values of the measurement covariance 
matrix to deal with numerical differentiation, resulted in the KF suppressing spikes of $\ddot{\textbf{m}}_{c,z}$ occurring at every foot 
strike (support exchange). This is also visible in \figref{estimation_accuracy}.

\subsection{Feedforward Compensation}

\begin{figure}[!t]
\parbox{\linewidth}{\centering
\includegraphics[width=0.49\linewidth]{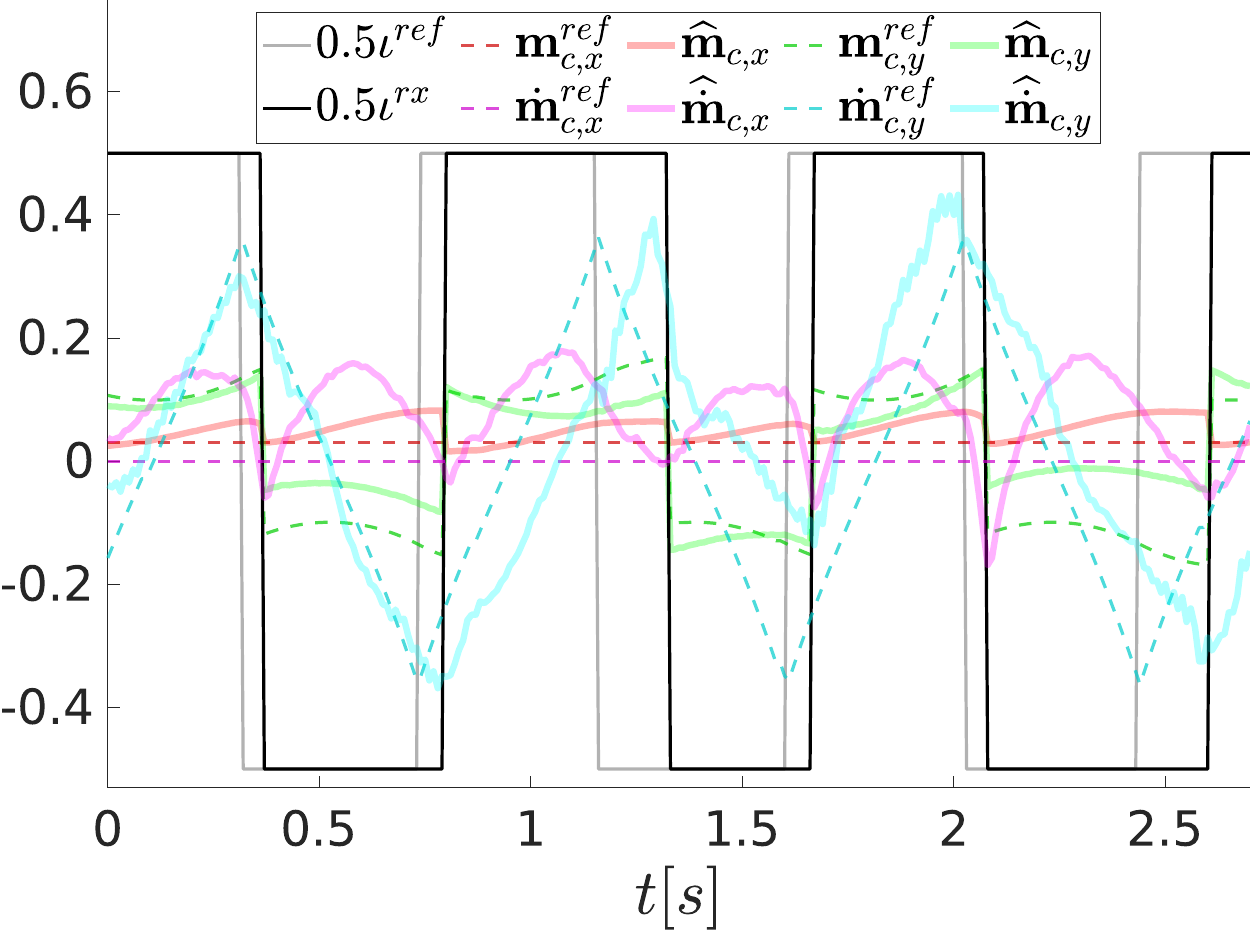}
\includegraphics[width=0.49\linewidth]{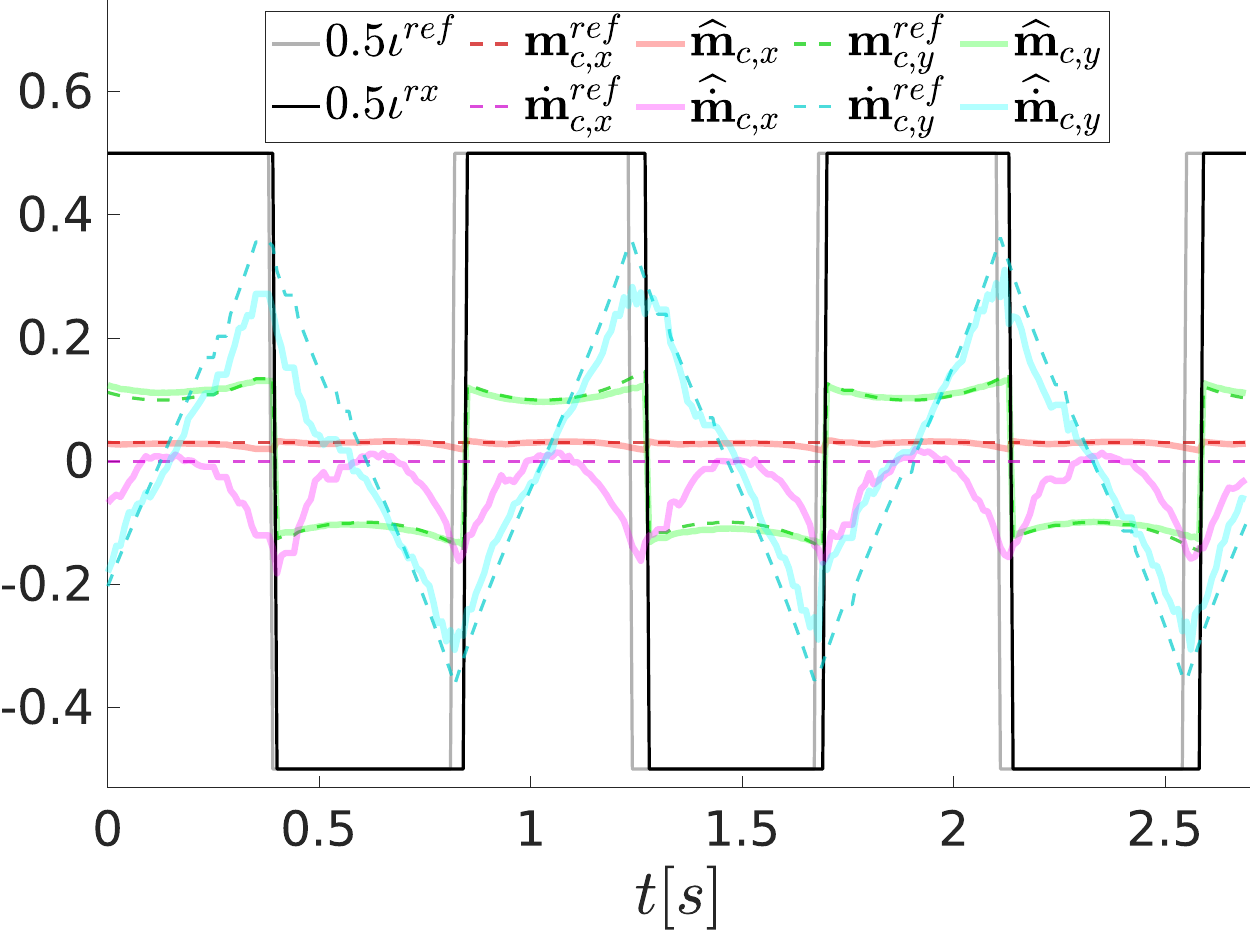}}
\caption{Effect of the feedforward compensation on the robots CoM state. Without the compensation (left), the robot is unable to maintain the desired nominal trajectories and keeps walking forward. With each step taken, the feet are placed at the wrong time and place. This consequently places the CoM on an undesired trajectory, disallowing the robot to get into a cyclic gait. After estimating and applying the feedforward terms (right), the feet are offset to correct for the errors, allowing the robot to follow the nominal state.}
\figlabel{feedforward}
\end{figure} 

We demonstrate the output produced by the developed compensation mechanisms in \figref{feedforward}, where in both cases the robot is asked to walk 
\textit{on the spot}. Without the feedforward components, the robot is unable to get into, or maintain a uniform walking rhythm. The timing, duration 
and sizes of the steps are all completely off, due to the dynamically changing load on the joints. In the sagittal direction the robot consistently 
keeps walking forward, as if bending under its own weight. 
This is an issue that the CPG-based gaits previously used on the NimbRo-OP2X did not face, as they are first and foremost meticulously tuned by hand 
to achieve open-loop walking. Once the estimation of the feedforward terms has been enabled, it takes only a few steps for the robot to completely 
correct its state. If left undisturbed, it can maintain it without issues. The latency $t_l$ of the system was estimated to be around \SI{55}{ms}. 
Although the tracking precision is already quite high, the estimated parameters can be further fine-tuned to achieve higher precision. The small 
errors can also be left for the feedback controller, which has been significantly relieved in its duties. 
Through several experiments we observed that the coupled nature of the sagittal and lateral planes necessitates correcting for errors in both simultaneously. 
This is also true for the coupling between step placement, its timing and angle of attack. All of these factors complement each other and intricately 
influence the state tracking and visual gait quality, which is not faithfully represented through ablation tests.

 \begin{figure*}[!t]
\parbox{\linewidth}{\centering
\includegraphics[height=0.30\linewidth]{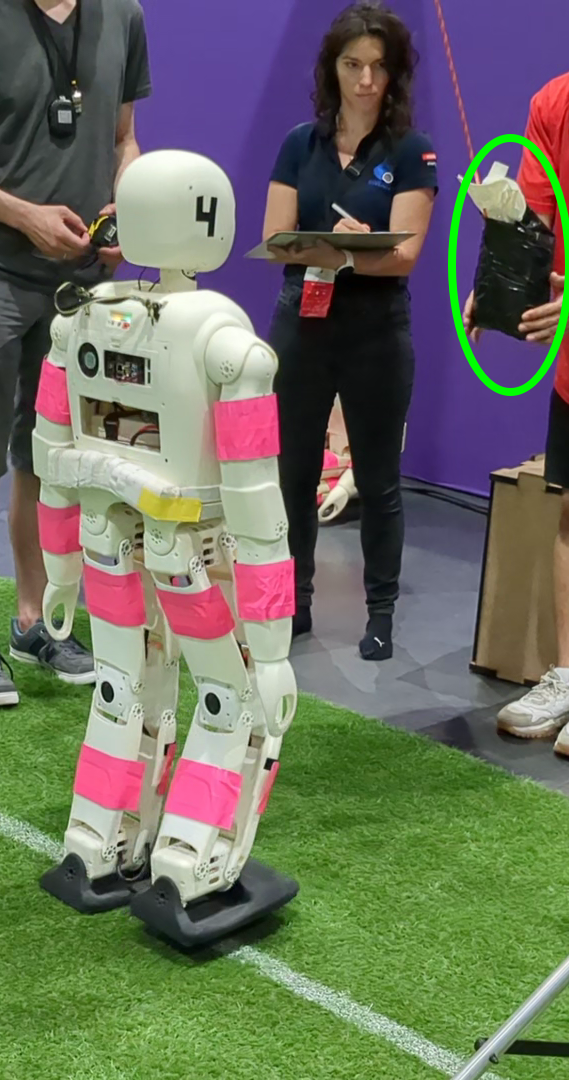}
\includegraphics[height=0.30\linewidth]{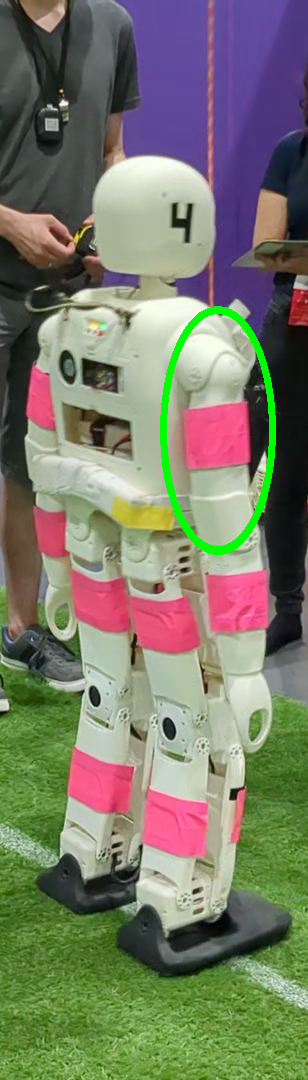}
\includegraphics[height=0.30\linewidth]{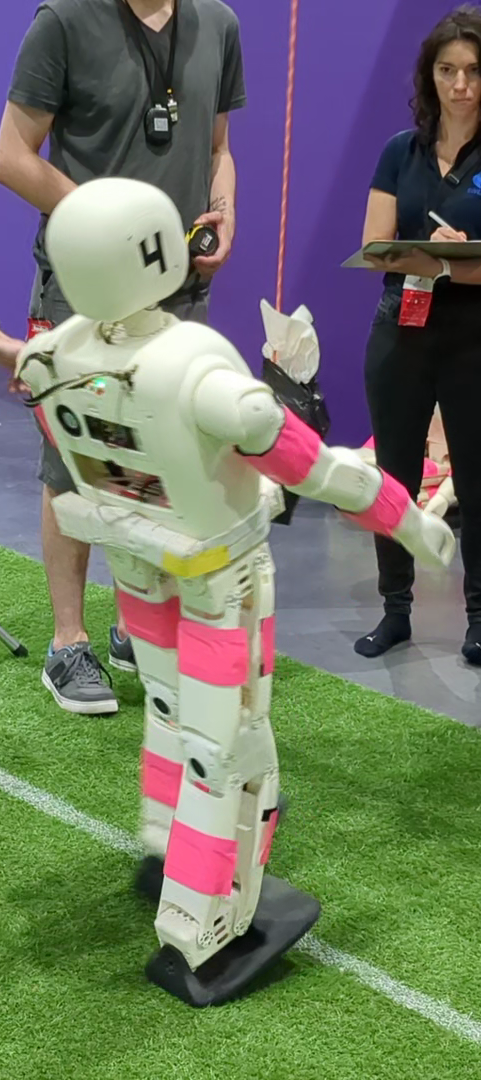}
\includegraphics[height=0.30\linewidth]{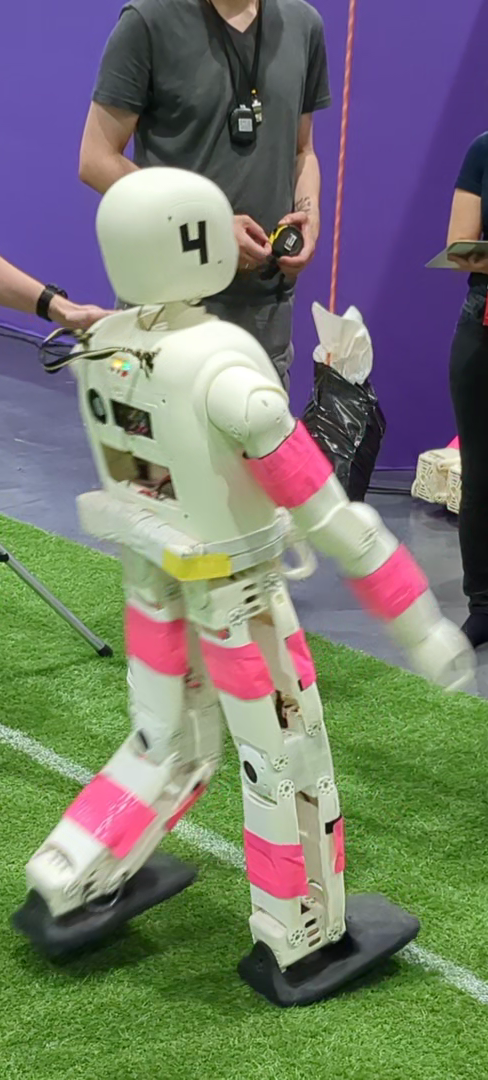}
\includegraphics[height=0.30\linewidth]{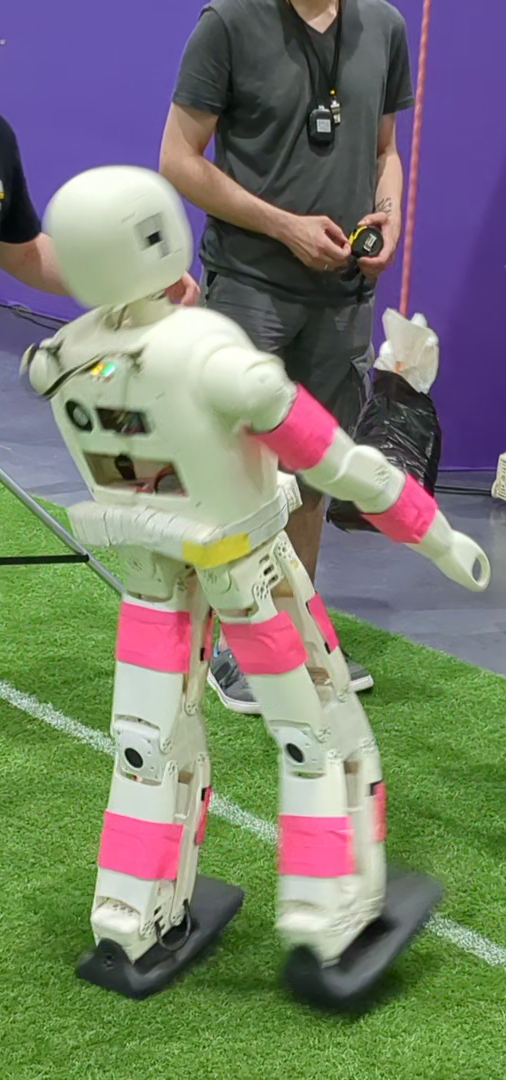}
\includegraphics[height=0.30\linewidth]{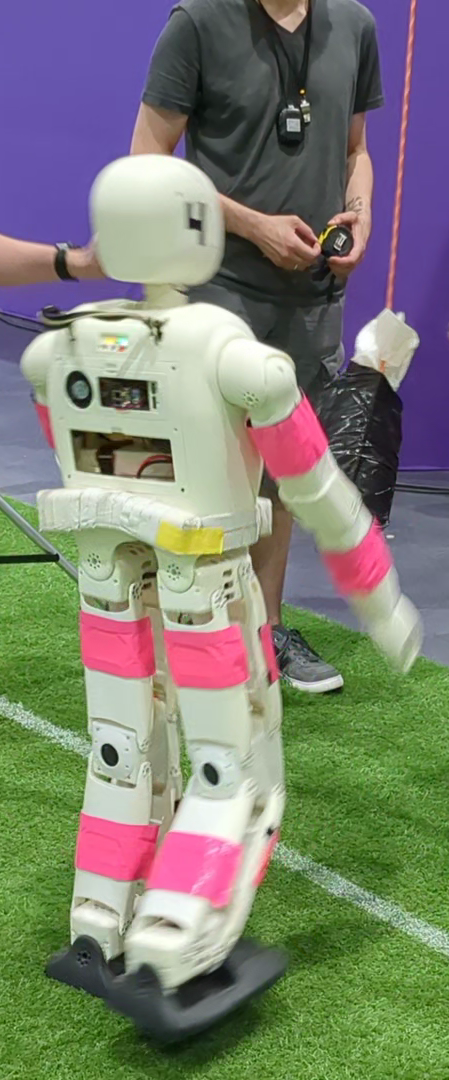}
\includegraphics[height=0.30\linewidth]{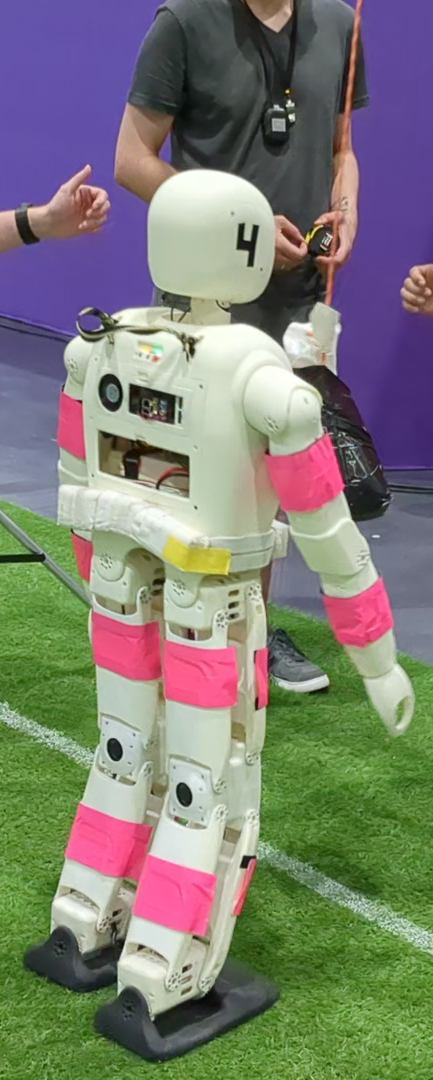}}
\caption{Still frames from the robot performing push recovery at RoboCup. The \SI{5}{kg} pendulum is circled in green. The first frame depicts the distance from which it was dropped, while the second one shows the moment of impact. After the impact, the robot undergoes an immediate and significant tilt. It manages to right itself using the developed reactive stepping approach.} 
\figlabel{pushrecovery}
\end{figure*}

\subsection{Balance}

The final component that we test is the balancing capability, which utilizes the presented approach in its entirety. 
The push-recovery capabilities have been presented during the Technical Challenges of RoboCup 2023, which was held in Bordeaux, France. 
During the challenge the robot has to complete a trial consisting of three consecutive pushes to its CoM, induced by a free-falling pendulum. 
One from the front, one from the back, and the third from either of the two. Using our approach, NimbRo-OP2X was able to successfully complete 
several trials with a \SI{3}{kg} and \SI{5}{kg} pendulum. The strongest pushes that the robot sustained were with a mass of \SI{5}{kg} which was let go 
at a horizontal distance of \SI{90}{cm} to the robot. An example of this trial is shown in \figref{pushrecovery}. The impact quickly starts to 
noticeably tilt the robot, which is met with an immediate response from the developed controllers. Firstly, the CoM controller quickly responds 
by shifting the weight of the robot towards the front, along with the arms. Simultaneously, the swing leg gets repositioned in accordance to the 
tilt and recomputed step position. The robot then continues walking until it fully disperses the push. Overall, this was the best performance in the 
2023 push-recovery challenge in the AdultSize class, more than doubling the score of the second team. This result also beats our team's previous 
performance achieved with the Capture Steps~\cite{missura2019capture} framework,  which in the 2019 and 2022 competitions was able to sustain 
perturbations of the \SI{5}{kg} pendulum from a distance of \SI{60}{cm}~\cite{pavlichenko2022robocup}. 
\begin{table}[b]
\renewcommand{\arraystretch}{1.02}
\renewcommand{\tabcolsep}{1.5mm}
\caption{Number of Withstood Simulated Pushes out of 10}
\tablabel{pushnumber}
\centering
\small
\begin{tabular}{c c c c c c c c c}
\hline

\hline
\bf{Force} [N] & from & 50  & 60  & 70  & 75  & 80  & 85 & 90 \\
\hline
\hline
\multirow{2}{*}{\bf{Presented framework}} & back  & 10  & 10 &  7 &  4  &  3  &  0 & 0  \\   
                         &  front  & 10  & 10   &  9 &  9  &  7  &  6 & 0  \\ 

\hline
\hline
\multirow{2}{*}{\bf{Capture steps~\cite{missura2019capture}}} & back & 10  & 10  & 4  & 0  & 0  & 0 & 0 \\
                                                        & front & 10  & 2  & 0  & 0  & 0  & 0 & 0 \\

\hline
\end{tabular}
\tablabel{pushtable}
\end{table}  

We also compare our approach to the mentioned Capture Steps framework quantitatively in simulation, as it was known to produce the best push-recovery 
results up to date with the NimbRo-OP2X robots. 
Both approaches were tuned to achieve the highest capable performance for the evaluation. The experiments 
consist of a series of pushes from the back and front, where our interface to MuJoCo applies the force to the center of the torso link for a duration of 
\SI{200}{ms}. After 10 pushes from both directions, the force gradually increases. Results of these experiments are summarized in \tabref{pushtable}. 
Both approaches can tolerate pushes of up to \SI{60}{N}. Above that, Capture Steps can only regain balance when pushed towards the front, without being 
able to recover a single push of \SI{75}{N}. This is most likely due to the noise suppression, adaptation and model blending factors, all of which just 
contribute to a slower response. The framework most likely would perform better for similar impulses, but having smaller forces spread across a longer 
time period. This is not a limitation in our approach, as the dynamics are accurately estimated and accounted for. This allows to tolerate even some 
pushes with a force of \SI{85}{N}. Opposite to Capture Steps, the pushes towards the back are tolerated better, due to the kinematics. As the robot 
cannot overextend its knee, taking a step back mostly moves the shank, while going forward requires to move the thigh (which is carrying the knee and 
the shank). At forces of \SI{90}{N} neither approach can recover, as the robot tilts and falls to the ground instantly.

\section{Conclusions}

We presented an approach to bipedal locomotion and balancing, specifically targeting platforms constrained by their hardware. The limitations include lack of sensing, imperfect actuation or even kinematic capabilities. We have quantitatively and qualitatively shown, that hardware with such limitations can also achieve relatively high performance. For this, we developed a novel state estimator based on a five-mass model with limb dynamics. The precise estimates were then used to establish a feedforward compensation scheme, operating on the centroidal state as opposed to the usual joint-space schemes. Finally, step feedback was extended to account for tilting due to kinematic limitations. 

In future work, we would like to investigate a meaningful way to combine both sagittal and lateral movement. The lateral movement in this work was mostly handled by pure CoM-ZMP and gain-based timing control. The reality is that the timing is much more nuanced and we hope to properly integrate  it with CAM regulation for a robust gait. We also want to explore possible improvements to the state estimator, which would incorporate joint current (torque) readings for more refined acceleration estimates.

\balance
\bibliographystyle{IEEEtran}
\bibliography{pcc}

\begin{thebibliography}{10}
\providecommand{\url}[1]{#1}
\csname url@samestyle\endcsname
\providecommand{\newblock}{\relax}
\providecommand{\bibinfo}[2]{#2}
\providecommand{\BIBentrySTDinterwordspacing}{\spaceskip=0pt\relax}
\providecommand{\BIBentryALTinterwordstretchfactor}{4}
\providecommand{\BIBentryALTinterwordspacing}{\spaceskip=\fontdimen2\font plus
\BIBentryALTinterwordstretchfactor\fontdimen3\font minus
  \fontdimen4\font\relax}
\providecommand{\BIBforeignlanguage}[2]{{%
\expandafter\ifx\csname l@#1\endcsname\relax
\typeout{** WARNING: IEEEtran.bst: No hyphenation pattern has been}%
\typeout{** loaded for the language `#1'. Using the pattern for}%
\typeout{** the default language instead.}%
\else
\language=\csname l@#1\endcsname
\fi
#2}}
\providecommand{\BIBdecl}{\relax}
\BIBdecl

\bibitem{al2012trajectory}
M.~Al~Borno, M.~De~Lasa, and A.~Hertzmann, ``Trajectory optimization for
  full-body movements with complex contacts,'' \emph{IEEE Transactions on
  Visualization and Computer Graphics}, vol.~19, no.~8, pp. 1405--1414, 2012.

\bibitem{lengagne2013generation}
S.~Lengagne, J.~Vaillant, E.~Yoshida, and A.~Kheddar, ``Generation of
  whole-body optimal dynamic multi-contact motions,'' \emph{The International
  Journal of Robotics Research}, vol.~32, no. 9-10, pp. 1104--1119, 2013.

\bibitem{kajita2010biped}
S.~Kajita, M.~Morisawa, K.~Miura, S.~Nakaoka, K.~Harada, K.~Kaneko,
  F.~Kanehiro, and K.~Yokoi, ``Biped walking stabilization based on linear
  inverted pendulum tracking,'' in \emph{IEEE/RSJ International Conference on
  Intelligent Robots and Systems (IROS)}, 2010, pp. 4489--4496.

\bibitem{dimitrov2011sparse}
D.~Dimitrov, A.~Sherikov, and P.-B. Wieber, ``A sparse model predictive control
  formulation for walking motion generation,'' in \emph{IEEE/RSJ International
  Conference on Intelligent Robots and Systems (IROS)}, 2011, pp. 2292--2299.

\bibitem{tedrake2015closed}
R.~Tedrake, S.~Kuindersma, R.~Deits, and K.~Miura, ``A closed-form solution for
  real-time zmp gait generation and feedback stabilization,'' in \emph{IEEE-RAS
  International Conference on Humanoid Robots (Humanoids)}, 2015, pp. 936--940.

\bibitem{pratt2006capture}
J.~Pratt, J.~Carff, S.~Drakunov, and A.~Goswami, ``Capture point: A step toward
  humanoid push recovery,'' in \emph{IEEE-RAS International Conference on
  Humanoid Robots (Humanoids)}, 2006, pp. 200--207.

\bibitem{englsberger2011bipedal}
J.~Englsberger, C.~Ott, M.~A. Roa, A.~Albu-Sch{\"a}ffer, and G.~Hirzinger,
  ``Bipedal walking control based on capture point dynamics,'' in
  \emph{IEEE/RSJ International Conference on Intelligent Robots and Systems
  (IROS)}, 2011, pp. 4420--4427.

\bibitem{englsberger2015three}
J.~Englsberger, C.~Ott, and A.~Albu-Sch{\"a}ffer, ``Three-dimensional bipedal
  walking control based on divergent component of motion,'' \emph{{IEEE}
  Transactions on Robotics}, vol.~31, no.~2, pp. 355--368, 2015.

\bibitem{orin2013centroidal}
D.~E. Orin, A.~Goswami, and S.-H. Lee, ``Centroidal dynamics of a humanoid
  robot,'' \emph{Autonomous Robots}, vol.~35, no. 2-3, pp. 161--176, 2013.

\bibitem{dai2014whole}
H.~Dai, A.~Valenzuela, and R.~Tedrake, ``Whole-body {M}otion {P}lanning with
  {C}entroidal {D}ynamics and {F}ull {K}inematics,'' in \emph{IEEE-RAS
  International Conference on Humanoid Robots (Humanoids)}, 2014.

\bibitem{chen2022angular}
Y.-M. Chen, G.~Nelson, R.~Griffin, M.~Posa, and J.~Pratt, ``Integrable
  whole-body orientation coordinates for legged robots,'' \emph{IEEE/RSJ
  International Conference on Intelligent Robots and Systems (IROS)}, 2023.

\bibitem{schuller2021online}
R.~Schuller, G.~Mesesan, J.~Englsberger, J.~Lee, and C.~Ott, ``Online
  centroidal angular momentum reference generation and motion optimization for
  humanoid push recovery,'' \emph{IEEE Robotics and Automation Letters (RAL)},
  vol.~6, no.~3, pp. 5689--5696, 2021.

\bibitem{ficht2023direct}
G.~Ficht and S.~Behnke, ``{D}irect {C}entroidal {C}ontrol for {B}alanced
  {H}umanoid {L}ocomotion,'' in \emph{Climbing and Walking Robots Conference
  (CLAWAR)}.\hskip 1em plus 0.5em minus 0.4em\relax Springer, 2023, pp.
  242--255.

\bibitem{ficht2018nimbro}
G.~Ficht, H.~Farazi, A.~Brandenburger, D.~Rodriguez, D.~Pavlichenko,
  P.~Allgeuer, M.~Hosseini, and S.~Behnke, ``Nimb{R}o-{OP2X}: Adult-sized
  open-source {3D} printed humanoid robot,'' in \emph{IEEE-RAS International
  Conference on Humanoid Robots (Humanoids)}.\hskip 1em plus 0.5em minus
  0.4em\relax IEEE, 2018, pp. 1--9.

\bibitem{Allgeuer2016a}
P.~Allgeuer and S.~Behnke, ``Omnidirectional bipedal walking with direct fused
  angle feedback mechanisms,'' in \emph{16th IEEE-RAS International Conference
  on Humanoid Robots (Humanoids)}, Canc\'un, Mexico, 2016.

\bibitem{missura2019capture}
M.~Missura, M.~Bennewitz, and S.~Behnke, ``Capture steps: Robust walking for
  humanoid robots,'' \emph{International Journal of Humanoid Robotics (IJHR)},
  vol.~16, no.~06, p. 1950032, 2019.

\bibitem{Schwarz2013a}
M.~Schwarz and S.~Behnke, ``Compliant robot behavior using servo actuator
  models identified by iterative learning control,'' in \emph{RoboCup
  International Symposium}, Eindhoven, Netherlands, 2013.

\bibitem{gehring2016practice}
C.~Gehring, S.~Coros, M.~Hutter, C.~D. Bellicoso, H.~Heijnen, R.~Diethelm,
  M.~Bloesch, P.~Fankhauser, J.~Hwangbo, M.~Hoepflinger \emph{et~al.},
  ``Practice makes perfect: An optimization-based approach to controlling agile
  motions for a quadruped robot,'' \emph{IEEE Robotics \& Automation Magazine},
  vol.~23, no.~1, pp. 34--43, 2016.

\bibitem{hwangbo2019learning}
J.~Hwangbo, J.~Lee, A.~Dosovitskiy, D.~Bellicoso, V.~Tsounis, V.~Koltun, and
  M.~Hutter, ``Learning agile and dynamic motor skills for legged robots,''
  \emph{Science Robotics}, vol.~4, no.~26, 2019.

\bibitem{wensing2017proprioceptive}
P.~M. Wensing, A.~Wang, S.~Seok, D.~Otten, J.~Lang, and S.~Kim,
  ``Proprioceptive actuator design in the {MIT C}heetah: Impact mitigation and
  high-bandwidth physical interaction for dynamic legged robots,'' \emph{IEEE
  Transactions on Robotics}, vol.~33, no.~3, pp. 509--522, 2017.

\bibitem{chignoli2021humanoid}
M.~Chignoli, D.~Kim, E.~Stanger-Jones, and S.~Kim, ``The {MIT} humanoid robot:
  Design, motion planning, and control for acrobatic behaviors,'' in
  \emph{IEEE-RAS International Conference on Humanoid Robots (Humanoids)},
  2021, pp. 1--8.

\bibitem{masuya2020review}
K.~Masuya and K.~Ayusawa, ``A review of state estimation of humanoid robot
  targeting the center of mass, base kinematics, and external wrench,''
  \emph{Advanced Robotics}, vol.~34, no. 21-22, pp. 1380--1389, 2020.

\bibitem{allgeuer2015fused}
P.~Allgeuer and S.~Behnke, ``Fused {A}ngles: A representation of body
  orientation for balance,'' in \emph{2015 IEEE/RSJ International Conference on
  Intelligent Robots and Systems (IROS)}, 2015, pp. 366--373.

\bibitem{ficht2020fast}
G.~Ficht and S.~Behnke, ``Fast whole-body motion control of humanoid robots
  with inertia constraints,'' in \emph{IEEE International Conference on
  Robotics and Automation (ICRA)}, 2020, pp. 6597--6603.

\bibitem{Allgeuer2014}
P.~Allgeuer and S.~Behnke, ``Robust sensor fusion for biped robot attitude
  estimation,'' in \emph{Proceedings of 14th IEEE-RAS Int. Conference on
  Humanoid Robotics (Humanoids)}, Madrid, Spain, 2014.

\bibitem{ludwig2018comparison}
S.~A. Ludwig and K.~D. Burnham, ``Comparison of {E}uler estimate using extended
  {K}alman {F}ilter, {M}adgwick and {M}ahony on quadcopter flight data,'' in
  \emph{International Conference on Unmanned Aircraft Systems (ICUAS)}, 2018,
  pp. 1236--1241.

\bibitem{cisneros2020lyapunov}
R.~Cisneros, A.~Benallegue, Y.~Chitour, M.~Morisawa, F.~Kanehiro \emph{et~al.},
  ``Lyapunov-stable orientation estimator for humanoid robots,'' \emph{IEEE
  Robotics and Automation Letters (RAL)}, vol.~5, no.~4, pp. 6371--6378, 2020.

\bibitem{herr2008angular}
H.~Herr and M.~Popovic, ``Angular momentum in human walking,'' \emph{Journal of
  Experimental Biology}, vol. 211, no.~4, pp. 467--481, 2008.

\bibitem{kajita20013d}
S.~Kajita, F.~Kanehiro, K.~Kaneko, K.~Yokoi, and H.~Hirukawa, ``The {3D} linear
  inverted pendulum mode: A simple modeling for a biped walking pattern
  generation,'' in \emph{IEEE/RSJ International Conference on Intelligent
  Robots and Systems (IROS)}, 2001, pp. 239--246.

\bibitem{kajita2019linear}
S.~Kajita, ``Linear inverted pendulum-based gait,'' \emph{Humanoid Robotics: A
  Reference}, pp. 905--922, 2019.

\bibitem{choi2007posture}
Y.~Choi, D.~Kim, Y.~Oh, and B.-J. You, ``Posture/walking control for humanoid
  robot based on kinematic resolution of com jacobian with embedded motion,''
  \emph{{IEEE} Transactions on Robotics}, vol.~23, no.~6, pp. 1285--1293, 2007.

\bibitem{pavlichenko2022robocup}
D.~Pavlichenko, G.~Ficht, A.~Amini, M.~Hosseini, R.~Memmesheimer,
  A.~Villar-Corrales, S.~M. Schulz, M.~Missura, M.~Bennewitz, and S.~Behnke,
  ``{RoboCup} 2022 {AdultSize} winner {NimbRo}: Upgraded perception, capture
  steps gait and phase-based in-walk kicks,'' in \emph{Robot World Cup}.\hskip
  1em plus 0.5em minus 0.4em\relax Springer, 2022, pp. 240--252.

\end{thebibliography}

\end{document}